\documentclass[11pt,letterpaper]{article}
\usepackage{naaclhlt2016}
\usepackage{times}
\usepackage{latexsym}
\usepackage{graphicx}
\usepackage{amsfonts}
\usepackage{verbatim}
\usepackage{amssymb}
\usepackage{parskip}
\usepackage{url}

\naaclfinalcopy 
\def\naaclpaperid{82} 

\title{MGNC-CNN: A Simple Approach to Exploiting\\ Multiple Word Embeddings for Sentence Classification}

\author{
    Ye Zhang$^1$ \hspace{.1\textwidth} {\bf Stephen Roller}$^1$\hspace{.1\textwidth} {\bf Byron C. Wallace}$^2$\\\\
    \begin{tabular}{cc}
        $^1${Department of Computer Science} & $^2$iSchool \\
        The University of Texas at Austin & The University of Texas at Austin\\
        \tt{\{yezhang,roller\}@cs.utexas.edu} & \tt{byron.wallace@utexas.edu}\\
    \end{tabular}
  }

\date{}

\begin{document}

\maketitle

\begin{abstract}
 We introduce a novel, simple convolution neural network (CNN) architecture -- \emph{multi-group norm constraint CNN (MGNC-CNN)} -- that capitalizes on multiple sets of word embeddings for sentence classification. MGNC-CNN extracts features from input embedding sets independently and then joins these at the penultimate layer in the network to form a final feature vector. We then adopt a group regularization strategy that differentially penalizes weights associated with the subcomponents generated from the respective embedding sets. This model is much simpler than comparable alternative architectures and requires substantially less training time. Furthermore, it is flexible in that it does not require input word embeddings to be of the same dimensionality. We show that MGNC-CNN consistently outperforms baseline models. 
 

\end{abstract}

\section{Introduction}
\label{section:intro}

Neural models have recently gained popularity for Natural Language Processing (NLP) tasks \cite{goldberg2015primer,collobert2008unified,cho2015natural}. For sentence classification, in particular, Convolution Neural Networks (CNN) have realized impressive performance \cite{kim2014convolutional,zhang2015sensitivity}. These models operate over \emph{word embeddings}, i.e., dense, low dimensional vector representations of words that aim to capture salient semantic and syntactic properties \cite{collobert2008unified}. 

An important consideration for such models is the specification of the word embeddings. Several options exist. For example, Kalchbrenner et al. \shortcite{kalchbrenner2014convolutional} initialize word vectors to random low-dimensional vectors to be fit during training, while Johnson and Zhang \shortcite{johnson2014effective} use fixed, one-hot encodings for each word. By contrast, Kim \shortcite{kim2014convolutional} initializes word vectors to those estimated via the word2vec model trained on 100 billion words of Google News~\cite{mikolov2013distributed}; these are then updated during training. Initializing embeddings to pre-trained word vectors is intuitively appealing because it allows transfer of learned distributional semantics. This has allowed a relatively simple CNN architecture to achieve remarkably strong results.

Many pre-trained word embeddings are now readily available on the web, induced using different models, corpora, and processing steps. Different embeddings may encode different aspects of language \cite{pado:cl07,erk:emnlp08,levy:acl14}: those based on bag-of-words (BoW) statistics tend to capture associations (\emph{doctor} and \emph{hospital}), while embeddings based on dependency-parses encode similarity in terms of use (\emph{doctor} and \emph{surgeon}). It is natural to consider how these embeddings might be \emph{combined} to improve NLP models in general and CNNs in particular. 

\noindent {\bf Contributions.} We propose MGNC-CNN, a novel, simple, scalable CNN architecture that can accommodate multiple off-the-shelf embeddings of variable sizes. Our model treats different word embeddings as distinct groups, and applies CNNs independently to each, thus generating corresponding feature vectors (one per embedding) which are then concatenated at the classification layer. Inspired by prior work exploiting regularization to encode structure for NLP tasks \cite{yogatama2014making,wallace-choe-charniak:2015:ACL-IJCNLP}, we impose different regularization penalties on weights for features generated from the respective word embedding sets.

Our approach enjoys the following advantages compared to the only existing comparable model \cite{yin-schutze:2015:CoNLL}: (i) It can leverage diverse, readily available word embeddings with different dimensions, thus providing flexibility. (ii) It is comparatively simple, and does not, for example, require \emph{mutual learning} or \emph{pre-training}.
(iii) It is an order of magnitude more efficient in terms of training time. 
\vspace{-.5em}
\section{Related Work} 
\vspace{-.5em}
\label{section:related} 



Prior work has considered combining latent representations of words that capture syntactic and semantic properties \cite{van2011latent}, and inducing multi-modal embeddings \cite{bruni2012distributional} for general NLP tasks. And recently, Luo et al. \shortcite{luo2014pre} proposed a framework that combines multiple word embeddings to measure text similarity, however their focus was not on classification. 


More similar to our work, Yin and Sch\"{u}tze \shortcite{yin-schutze:2015:CoNLL} proposed \emph{MVCNN} for sentence classification. This CNN-based architecture accepts multiple word embeddings as inputs. These are then treated as separate `channels', analogous to RGB channels in images. Filters consider all channels simultaneously. MVCNN achieved state-of-the-art performance on multiple sentence classification tasks. However, this model has practical drawbacks. (i) MVCNN requires that input word embeddings have the same dimensionality. Thus to incorporate a second set of word vectors trained on a corpus (or using a model) of interest, one needs to either find embeddings that happen to have a set number of dimensions or to estimate embeddings from scratch. (ii) The model is complex, both in terms of implementation and run-time. Indeed, this model requires pre-training and mutual-learning and requires days of training time, whereas the simple architecture we propose requires on the order of an hour (and is easy to implement).

\section{Model Description}
\label{section:model} 

We first review standard one-layer CNN (which exploits a single set of embeddings) for sentence classification \cite{kim2014convolutional}, and then propose our augmentations, which exploit multiple embedding sets. 

\textbf{Basic CNN.} In this model we first replace each word in a sentence with its vector representation, resulting in a sentence matrix $\mathbf{A}\in \mathbb{R}^{s\times d}$, where $s$ is the (zero-padded) sentence length, and $d$ is the dimensionality of the embeddings. We apply a convolution operation between linear filters with parameters $\mathbf{w}_1,\mathbf{w}_2,...,\mathbf{w}_k$ and the sentence matrix. For each $\mathbf{w}_i\in \mathbb{R}^{h\times d}$, where $h$ denotes `height', we slide filter $i$ across $\mathbf{A}$, considering `local regions' of $h$ adjacent rows at a time.  
At each local region, we perform element-wise multiplication and then take the element-wise sum between the filter and the (flattened) sub-matrix of $\mathbf{A}$, producing a scalar. We do this for each sub-region of $\mathbf{A}$ that the filter spans, resulting in a \emph{feature map} vector $\mathbf{c}_i\in \mathbb{R}^{(s-h+1) \times 1}$. We can use multiple filter sizes with different heights, and for each filter size we can have multiple filters. Thus the model comprises $k$ weight vectors $\mathbf{w}_1, \mathbf{w}_2, ... \mathbf{w}_k$, each of which is associated with an instantiation of a specific filter size. These in turn generate corresponding feature maps $\mathbf{c}_1,\mathbf{c}_2,...\mathbf{c}_k$, with dimensions varying with filter size. A \emph{1-max pooling} operation is applied to each feature map, extracting the largest number $o_i$ from each feature map $i$. Finally, we combine all $o_i$ together to form a feature vector $\mathbf{o}\in \mathbb{R}^k$ to be fed through a softmax function for classification. We regularize weights at this level in two ways. (1) \emph{Dropout}, in which we randomly set elements in $\mathbf{o}$ to zero during the training phase with probability $p$, and multiply $p$ with the parameters trained in $\mathbf{o}$ at test time. (2) An l2 norm penalty, for which we set a threshold $\lambda$ for the l2 norm of $\mathbf{o}$ during training; if this is exceeded, we rescale the vector accordingly. For more details, see \cite{zhang2015sensitivity}.

\begin{figure}[h]
\includegraphics[width=0.48\textwidth]{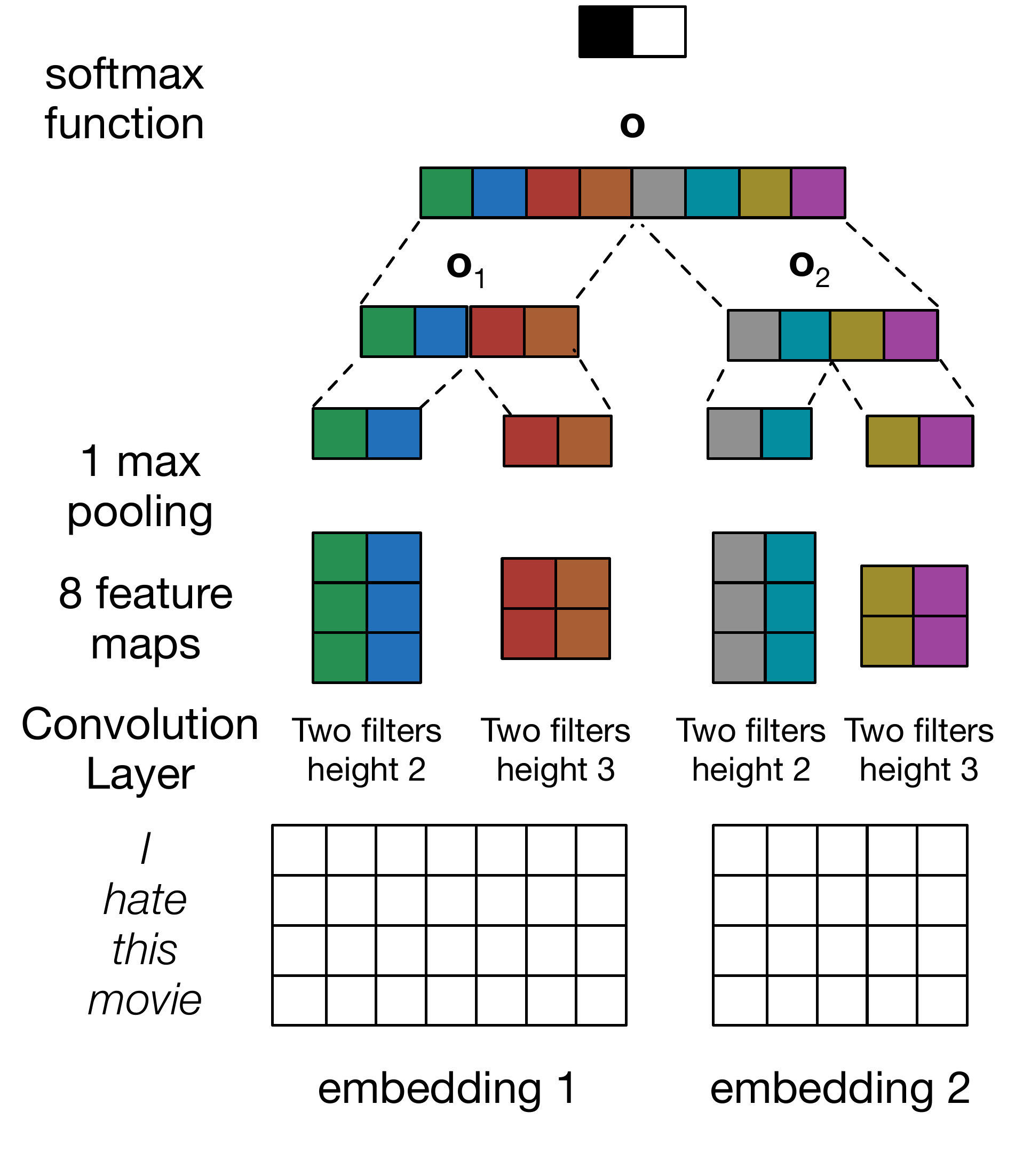}
\caption{Illustration of MG-CNN and MGNC-CNN. The filters applied to the respective embeddings are completely independent. MG-CNN applies a max norm constraint to $\mathbf{o}$, while MGNC-CNN applies max norm constraints on $\mathbf{o_1}$ and $\mathbf{o_2}$ independently (group regularization). Note that one may easily extend the approach to handle more than two embeddings at once.}
\label{fig:MG-CNN}
\end{figure}

\textbf{MG-CNN.} Assuming we have $m$ word embeddings with corresponding dimensions $d_1,d_2,...d_m$, we can simply treat each word embedding independently. In this case, the input to the CNN comprises multiple sentence matrices $\mathbf{A_1},\mathbf{A_2},...\mathbf{A_m}$, where each $\mathbf{A}_l \in \mathbb{R}^{s\times d_l}$ may have its own width $d_l$. We then apply different groups of filters $\{\mathbf{w}_1\},\{\mathbf{w}_2\},...\{\mathbf{w}_m\}$ independently to each $\mathbf{A}_l$, where $\{\mathbf{w}_l\}$ denotes the set of filters for $\mathbf{A}_l$. As in basic CNN, $\{\mathbf{w}_l\}$ may have multiple filter sizes, and multiple filters of each size may be introduced. At the classification layer we then obtain a feature vector $\mathbf{o}_l$ for each embedding set, and we can simply concatenate these together to form the final feature vector $\mathbf{o}$ to feed into the softmax function, where $\mathbf{o}=\mathbf{o_1}\oplus\mathbf{o_2}...\oplus\mathbf{o_m}$. This representation contains feature vectors generated from all sets of embeddings under consideration. We call this method \emph{multiple group CNN} (MG-CNN). Here groups refer to the features generated from different embeddings. Note that this differs from `multi-channel' models because at the convolution layer we use different filters on each word embedding matrix independently, whereas in a standard multi-channel approach each filter would consider all channels simultaneously and generate a scalar from all channels at each local region. As above, we impose a max l2 norm constraint on the final feature vector $\mathbf{o}$ for regularization. Figure \ref{fig:MG-CNN} illustrates this approach.


\textbf{MGNC-CNN}. We propose an augmentation of MG-CNN, \emph{Multi-Group Norm Constraint CNN} (MGNC-CNN), which differs in its regularization strategy. Specifically, in this variant we impose \emph{grouped} regularization constraints, independently regularizing subcomponents $\mathbf{o}_l$ derived 
from the respective embeddings, i.e., we impose separate max norm constraints $\lambda_l$ for each $\mathbf{o}_l$ (where $l$ again indexes embedding sets); these $\lambda_l$ hyper-parameters are to be tuned on a validation set. Intuitively, this method aims to better capitalize on features derived from word embeddings that capture discriminative properties of text for the task at hand by penalizing larger weight estimates for features derived from less discriminative embeddings. 


\section{Experiments}

\begin{table*}[t]
\scriptsize
\centering
\begin{tabular}{c c c c c c} 
 \emph{Model} & \emph{Subj} & \emph{SST-1} & \emph{SST-2} & \emph{TREC} & \emph{Irony} \\ 
 \hline
 CNN(w2v) & 93.14 (92.92,93.39) &46.99 (46.11,48.28) &87.03 (86.16,88.08) & 93.32 (92.40,94.60)& 67.15 (66.53,68.11)\\ 
 CNN(Glv) &93.41(93.20,93.51)   &46.58 (46.11,47.06) &87.36 (87.20,87.64) & 93.36 (93.30,93.60) & 67.84 (67.29,68.38)\\ 
 CNN(Syn) &93.24(93.01,93.45) &45.48(44.67,46.24) & 86.04 (85.28,86.77) & 94.68 (94.00,95.00) & 67.93 (67.30,68.38)\\
 MVCNN~\cite{yin-schutze:2015:CoNLL}& 93.9 & 49.6 & 89.4 & - & -\\
 C-CNN(w2v+Glv)& 93.72 (93.68,93.76) & 47.02(46.24,47.69) & 87.42(86.88,87.81) & 93.80 (93.40,94.20) & 67.70 (66.97,68.35)\\
 C-CNN(w2v+Syn) &93.48 (93.43,93.52) &46.91(45.97,47.81) & 87.17 (86.55,87.42) &94.66 (94.00,95.20) & 68.08 (67.33,68.57)\\ 
 C-CNN(w2v+Syn+Glv)& 93.61 (93.47,93.77) & 46.52 (45.02,47.47) & 87.55 (86.77,88.58) & 95.20 (94.80,65.60)& 68.38 (67.66,69.23)\\
 MG-CNN(w2v+Glv) &93.84 (93.66,94.35) &48.24 (47.60,49.05) & 87.90 (87.48,88.30)& 94.09 (93.60,94.80) & 69.40 (66.35,72.30)\\
 MG-CNN(w2v+Syn) &93.78 (93.62,93.98)&48.48(47.78,49.19) &87.47(87.10,87.70) & 94.87 (94.00,95.60) & 68.28 (66.44,69.97)\\
 MG-CNN(w2v+Syn+Glv) & \textbf{94.11 (94.04,94.17)} & 48.01 (47.65,48.37)&87.63(87.04,88.36) & 94.68 (93.80,95.40) & 69.19(67.06,72.30)\\
 MGNC-CNN(w2v+Glv) &93.93 (93.79,94.14) &48.53 (47.92,49.37) & \textbf{88.35(87.86,88.74)} & 94.40 (94.00,94.80) & 69.15 (67.25,71.70) \\
 MGNC-CNN(w2v+Syn) &93.95 (93.75,94.21) &48.51 (47.60,49.41)&87.88(87.64,88.19) & 95.12 (94.60,95.60) & 69.35 (67.40,70.86)\\
 MGNC-CNN(w2v+Syn+Glv) & 94.09 (93.98,94.18) & \textbf{48.65 (46.92,49.19)} & 88.30 (87.83,88.65) & \textbf{95.52 (94.60,96.60) }& \textbf{71.53 (69.74,73.06)}
\end{tabular}
\caption{Results mean (min, max) achieved with each method.
\label{table:results}
w2v:word2vec. Glv:GloVe. Syn: Syntactic embedding. Note that we experiment with using two and three sets of embeddings jointly, e.g., w2v+Syn+Glv indicates that we use all three of these.}

\vspace*{10 mm}

\scriptsize
\centering
\begin{tabular}{c c c c c c} 
 \emph{Model} & \emph{Subj} & \emph{SST-1} & \emph{SST-2} & \emph{TREC} & \emph{Irony} \\ 
 \hline
 CNN(w2v) & 9 &81 &81& 9 & 243\\ 
 CNN(Glv) &3 &9&1& 9 & 81 \\ 
 CNN(Syn) &3  &81& 9 & 81 & 1\\
 C-CNN(w2v+Glv)& 9 & 9 & 3 & 3 & 1\\
 C-CNN(w2v+Syn) & 3 &81 & 9&9 & 1\\ 
 C-CNN(w2v+Syn+Glv) & 9 & 9 & 1 & 81& 81\\
 MG-CNN(w2v+Glv) & 3 & 9 &3 & 81 & 9\\
 MG-CNN(w2v+Syn) & 9 &81 & 3 & 81 & 3\\
 MG-CNN(w2v+Syn+Glv) & 9 & 1 & 9 & 243 & 9 \\
 MGNC-CNN(w2v+Glv) & (9,3) &(81,9) &(1,1) & (9,81) & (243,243) \\
 MGNC-CNN(w2v+Syn) &(3,3) &(81,81) &(81,9) & (81,81) & (81,3)\\
 MGNC-CNN(w2v+Syn+Glv)& (81,81,81) & (81,81,1) & (9,9,9) & (1,81,81) & (243,243,3)\\
\end{tabular}
\caption{Best $\lambda^2$ value on the validation set for each method
\label{table:hyper}
w2v:word2vec. Glv:GloVe. Syn: Syntactic embedding.}
\end{table*}

\subsection{Datasets}
\textbf{Stanford Sentiment Treebank}
Stanford Sentiment Treebank (SST) \cite{socher2013recursive}.
This concerns predicting movie review sentiment. Two datasets are derived from this corpus: (1) {\bf SST-1}, containing five classes: \emph{very negative}, \emph{negative}, \emph{neutral}, \emph{positive}, and \emph{very positive}. (2) {\bf SST-2}, which has only two classes: \emph{negative} and \emph{positive}. For both, we remove phrases of length less than 4 from the training set.\footnote{As in \cite{kim2014convolutional}.}
\noindent \textbf{Subj} \cite{pang2004sentimental}. The aim here is to classify sentences as either \emph{subjective} or \emph{objective}. This comprises 5000 instances of each.
\noindent \textbf{TREC} \cite{li2002learning}. A question classification dataset containing six classes: \emph{abbreviation}, \emph{entity}, \emph{description}, \emph{human}, \emph{location} and \emph{numeric}. There are 5500 training and 500 test instances. 
\noindent \textbf{Irony} \cite{wallace2014humans}. This dataset
contains 16,006 sentences from reddit labeled as
ironic (or not). The dataset is imbalanced (relatively
few sentences are ironic). Thus before training,
we under-sampled negative instances to make
classes sizes equal. Note that for this dataset we report the
Area Under Curve (AUC), rather than accuracy,
because it is imbalanced.

\subsection{Pre-trained Word Embeddings}
We consider three sets of word embeddings for our experiments: (i) {\bf word2vec}\footnote{\url{https://code.google.com/p/word2vec/}} is trained on 100 billion tokens of Google News dataset; (ii) {\bf GloVe}~\cite{pennington2014glove}\footnote{\url{http://nlp.stanford.edu/projects/glove/}} is trained on aggregated global word-word co-occurrence statistics from Common Crawl (840B tokens); and (iii) {\bf syntactic} word embedding trained on dependency-parsed corpora. These three embedding sets happen to all be 300-dimensional, but our model could accommodate arbitrary and variable sizes.

We pre-trained our own syntactic embeddings following \cite{levy:acl14}. We parsed the ukWaC corpus \cite{baroni:lrec09} using the Stanford Dependency Parser v3.5.2 with Stanford Dependencies \cite{chen:emnlp14} and extracted (word, relation+context) pairs from parse trees. We ``collapsed" nodes with prepositions and notated inverse relations separately, e.g., ``dog barks" emits two tuples: {\em(barks, nsubj\_dog)} and  {\em (dog, nsubj$^{-1}$\_barks)}. We filter words and contexts that appear fewer than 100 times, resulting in $\sim$173k words and 1M contexts. We trained 300d vectors using word2vecf\footnote{\url{https://bitbucket.org/yoavgo/word2vecf/}} with default parameters.

\subsection{Setup}

We compared our proposed approaches to a standard CNN that exploits a single set of word embeddings \cite{kim2014convolutional}. We also compared to a baseline of simply concatenating embeddings for each word to form long vector inputs. We refer to this as Concatenation-CNN \emph{C-CNN}. For all multiple embedding approaches (C-CNN, MG-CNN and MGNC-CNN), we explored two combined sets of embedding: word2vec+Glove, and word2vec+syntactic, and one three sets of embedding: word2vec+Glove+syntactic. For all models, we tuned the l2 norm constraint $\lambda$ over the range $\{\frac{1}{3},1,3,9,81,243\}$ on a validation set. For instantiations of MGNC-CNN in which we exploited two embeddings, we tuned both $\lambda_1$, and $\lambda_2$; where we used three embedding sets, we tuned $\lambda_1,\lambda_2$ and $\lambda_3$.

We used standard train/test splits for those datasets that had them. Otherwise, we performed 10-fold cross validation, creating nested development sets with which to tune hyperparameters. For all experiments we used filters sizes of 3, 4 and 5 and we created 100 feature maps for each filter size. We applied 1 max-pooling and dropout (rate: 0.5) at the classification layer. For training we used back-propagation in mini-batches and used AdaDelta as the stochastic gradient descent (SGD) update rule, and set mini-batch size as 50. In this work, we treat word embeddings as part of the parameters of the model, and update them as well during training. In all our experiments, we only tuned the max norm constraint(s), fixing all other hyperparameters. 

\subsection{Results and Discussion}
\vspace{-.5em}
\label{section:results} 

We repeated each experiment 10 times and report the mean and ranges across these. This replication is important because training is stochastic and thus introduces variance in performance \cite{zhang2015sensitivity}. Results are shown in Table \ref{table:results}, and the corresponding best norm constraint value is shown in Table \ref{table:hyper}. We also show results on \emph{Subj}, \emph{SST-1} and \emph{SST-2} achieved by the more complex model of \cite{yin-schutze:2015:CoNLL} for comparison; this represents the state-of-the-art on the three datasets other than TREC. 


We can see that {\bf MGNC-CNN and MG-CNN always outperform baseline methods (including C-CNN), and MGNC-CNN is usually better than MG-CNN}.
And on the \emph{Subj} dataset, MG-CNN actually achieves slightly better results than \cite{yin-schutze:2015:CoNLL}, with far less
complexity and required training time (MGNC-CNN performs comparably, although no better, here). On the TREC dataset, the best-ever accuracy we are aware of is 96.0\% \cite{mou2015discriminative}, which falls within the range of the result of our MGNC-CNN model with three word embeddings. On the \emph{irony} dataset, our model with three embeddings achieves 4\% improvement (in terms of AUC) compared to the baseline model.  

On \emph{SST-1} and \emph{SST-2}, our model performs slightly worse than \cite{yin-schutze:2015:CoNLL}. However, we again note that their performance is achieved using a much more complex model which involves pre-training and mutual-learning steps. This model takes days to train, whereas our model requires on the order of an hour. 

We note that the method proposed by Astudillo \emph{et al.} ~\shortcite{astudillo2015learning} is able to accommodate multiple embedding sets with different dimensions by projecting the original word embeddings into a lower-dimensional space. However, this work requires training the optimal projection matrix on laebled data first, which again incurs large overhead. 


Of course, our model also has its own limitations: in MGNC-CNN, we need to tune the norm constraint hyperparameter for all the word embeddings. As the number of word embedding increases, this will increase the running time. However, this tuning procedure is embarrassingly parallel.

\vspace{-1em}
\section{Conclusions}
\vspace{-.5em}
We have proposed MGNC-CNN: a simple, flexible CNN architecture for sentence classification that can exploit multiple, variable sized word embeddings. We demonstrated that this consistently achieves better results than a baseline architecture that exploits only a single set of word embeddings, and also a naive concatenation approach to capitalizing on multiple embeddings. Furthermore, our results are comparable to those achieved with a recently proposed model \cite{yin-schutze:2015:CoNLL} that is much more complex. However, our simple model is easy to implement and requires an order of magnitude less training time. Furthermore, our model is much more flexible than previous approaches, because it can accommodate variable-size word embeddings. 

\vspace{-1em}
\section*{Acknowledgments}
\vspace{-.5em}

This work was supported in part by the Army Research Office (grant W911NF-14-1-0442) and by The Foundation for Science and Technology, Portugal (grant UTAP-EXPL/EEIESS/0031/2014). This work was also made possible by the support of the Texas Advanced Computer Center (TACC) at UT Austin.

\clearpage
\bibliography{naaclhlt2016}
\bibliographystyle{naaclhlt2016}

\end{document}